# A rapid approach to urban traffic noise mapping with a generative adversarial network

Xinhao Yang[1,2,†], Zhen Han[3,†], Xiaodong Lu[4], Yuan Zhang[1,2,*]

1 School of Architecture and Urban Planning, Shenyang Jianzhu University, Shenyang 110168, China

2 Liaoning Provincial Key Laboratory of Eco-Building Physics Technology and Evaluation, Shenyang Jianzhu University, Shenyang 110168, China

3 Department of Architecture, National University of Singapore, Singapore, 119077, Singapore

4 School of Architecture and Fine Art, Dalian University of Technology, Dalian 116023, China

† These authors contributed equally to this work.

* Correspondence to: 25 Hunnan Middle Road, Shenyang 110168, China. E-mail address: y.zhang@sjzu.edu.cn (Yuan Zhang).

Accepted by *Applied Acoustics*
Received 23 May 2024; Accepted 31 August 2024
DOI: 10.1016/j.apacoust.2024.110268

## Highlights

1. A rapid noise mapping model was developed based on a generative adversarial network
2. Noise prediction was enabled by identifying urban spatial features such as roads and buildings.
3. This model meets the rapid iteration needs of early urban planning design stages.
4. A tool was created by integration into Grasshopper to allow for casual use by non-acoustic experts.

## Abstract

With rapid urbanisation and the accompanying increase in traffic density, traffic noise has become a major concern in urban planning. However, traditional grid noise mapping methods have limitations in terms of time consumption, software costs, and a lack of parameter integration interfaces. These limitations hinder their ability to meet the need for iterative updates and rapid performance feedback in the early design stages of street-scale urban planning. Herein, we developed a rapid urban traffic noise mapping technique that leverages generative adversarial networks (GANs) as a surrogate model. This approach enables the rapid assessment of urban traffic noise distribution by using urban elements such as roads and buildings as the input. The mean values for the mean squared error (RMSE) and structural similarity index (SSIM) are 0.3024 dB(A) and 0.8528, respectively, for the validation dataset. The trained model is integrated into Grasshopper as a tool, facilitating the rapid generation of traffic noise maps. This integration allows urban designers and planners, even those without expertise in acoustics, to easily anticipate changes in acoustics impacts caused by design in the early design stages.

## 1 Introduction

Amidst rapid global urbanisation, enhancing the quality of urban life has emerged as a crucial objective of social development. A healthy and comfortable living environment is not only a fundamental requirement for residents but also a central focus of contemporary urban design. Among the various factors influencing the quality of life of urban areas, noise pollution has become an issue that cannot be overlooked. In addition to causing annoyance (Yang et al., 2021; Zhang et al., 2017), extensive epidemiological studies have demonstrated that individuals exposed to prolonged periods of high-intensity noise face an elevated risk of cardiovascular disease (Begou et al., 2020; Zhang et al., 2023), sleep disorders (Myllyntausta et al., 2020), hearing loss (Wang et al., 2021), and cognitive decline (Van Kempen et al., 2018).

The intensification of urbanisation is usually accompanied by an increase in the urban population. It is estimated that by 2030, 60% of the world's population will reside in urban areas (United Nations, 2015). This further catalysed the extensive construction of traffic road networks, resulting in traffic noise becoming a primary source of urban noise pollution (Lu et al., 2019). Urban residents living near roads are adversely affected by this noise. In China, the volume of traffic noise complaints has long been at the forefront of noise complaints (China, 2023). Furthermore, noise levels that exceed standard traffic thresholds can impact property prices in adjacent areas (Theebe, 2004). Consequently, the effective evaluation and control of traffic noise have emerged as shared concerns within the realms of public health and urban planning.

In 2002, the European Union (EU) promulgated the Environmental Noise Assessment and Management Directive, mandating that member countries regularly produce and revise noise maps to monitor environmental noise. Subsequently, noise maps have been extensively employed as a tool for the assessment and management of urban noise. Through the visual representation of noise levels, noise maps have become an effective tool for policy makers to develop appropriate policies and guide urban development (De Vos and Licitra, 2013), the related research involves the development and optimization of noise models (Alam et al., 2020; Ascigil-Dincer and Demirkale, 2021; Cook et al., 2023), simulation and evaluation (Ozkurt, 2014; Sadr et al., 2014; Vogiatzis, 2012), as well as the analysis of factors affecting noise distribution (Zhou et al., 2024).

Undoubtedly, considering the impact of traffic noise in the early design stages of urban planning is beneficial for reducing design and economic costs (De Coensel et al., 2010). Urban planning involves various stakeholders, such as government departments, planners, architects, developers, and consultants, each with different design requirements. This complexity makes urban planning a naturally multiobjective problem. Despite the increasing attention given to noise issues, noise issues are often overlooked in the early design stages during actual implementation processes. This oversight results in neglecting potential noise risks in design outcomes, leading to significant challenges in later design stages or requiring substantial economic investment after completion to address noise issues. Many researchers emphasise strengthening the direct relationship between design solutions and environmental performance to reduce unnecessary investments in later design stages (Huang et al., 2022).

However, traditional grid-based noise mapping methods and the noise map simulation software applications developed have limitations in the early design stages. These software programs are designed with noise assessment and management objectives and are primarily utilised for environmental impact assessments before finalising a design or for noise assessments on an urban scale post-project completion. In operation, they depend on well-established traffic networks and urban spatial planning schemes, necessitating intricate modelling processes and acoustic configurations for noise simulation tasks. This presents challenges for non-acoustic experts to refine and enhance solutions during the design phase. This approach not only demands a substantial amount of time but also entails significant financial costs to procure widely accepted acoustic calculation software (Gomez Escobar et al., 2012; Oiamo et al., 2018).

Moreover, a significant drawback of these existing systems is their lack of interfaces for parametric integration.

These limitations render them suboptimal for scenarios requiring extensive iterative updates during the preliminary phases of design, which necessitate rapid performance feedback, which is a common requirement in contemporary design processes (Li et al., 2024). During the early design stages, the architectural configurations of the surroundings and the structures of urban road networks are uncertain and prone to modifications. Therefore, there is an urgent need to develop new methods and tools to efficiently conduct noise simulations for rapid design iteration in the early stages of urban planning.

The advent of artificial intelligence techniques, particularly deep learning, has opened new possibilities on this front. Artificial neural networks, for example, offer scalable and adaptive solutions that may surmount the constraints inherent in grid-based approaches. In 2014, generative adversarial networks (GANs) proposed by Goodfellow drove the rapid development of image translation and image prediction (Goodfellow et al., 2014). A prevalent application of GANs is image-to-image translation, wherein one form of image data is transformed into another image data. In 2018, Isola et al. introduced the pix2pix algorithm, which is based on the conditional adversarial generative network (CGAN) model, to address the issue of image mapping (Isola et al., 2017). This algorithm has since gained widespread adoption in the realm of building performance prediction.

In the domain of architecture, deep learning technology has been widely applied in assisting design (Han et al., 2021b; Ko et al., 2023) and performance prediction (Singh et al., 2021; Singh et al., 2020). The application of GAN models facilitates the rapid prediction of simulation outcomes for indoor environments. This is particularly evident in the context of matrix-type data analysis, such as the assessment of interior lighting dynamics or the distribution of wind flow within built spaces (Han et al., 2021a; Mokhtar et al., 2020). This allows for the circumvention of extensive computational simulations. He et al. utilised the pix2pix algorithm to expedite the prediction of interior lighting scenarios in buildings (He et al., 2021). Hu et al. employed a multimodal GAN model to predict wind pressure patterns around buildings (Hu et al., 2020). These studies underscore the utility of GANs in the context of rapid building performance prediction.

Despite these advancements, there remains a paucity of research on the rapid prediction and visualisation of urban noise.

Building on this premise, the present study proposes a rapid urban noise mapping technique that uses a GAN as a surrogate model. This method enables rapid assessment of urban traffic noise distribution by accurately identifying urban elements such as roads and buildings within the input grid. After accuracy verification, the trained model is embedded as a tool in Grasshopper. This enables the tool to swiftly generate corresponding traffic noise images on the basis of the urban plan map provided by the user, aiding city designers and planners in assessing potential traffic noise hazards in the early design stages through the utilisation of predictive outcomes. This, in turn, facilitates the optimisation of solutions for urban spatial configurations and road networks.

## 2 Methods

In this study, a GAN-based surrogate model for urban noise map prediction was developed. The model training process is shown in Figure 1. First, the urban plane was selected from OpenStreetMap (OSM), and a database was generated by subdividing it into several regions using Grasshopper. Cadna/A, a widely used conventional outdoor ambient noise simulation software, was then employed to simulate traffic noise within the chosen urban areas. Grasshopper was used to visualise these simulations to create the training and validation datasets. The Pix2pix neural network was subsequently trained with this dataset for the rapid prediction of urban traffic noise. Finally, the performance of the model was evaluated against the pre-prepared validation dataset not included in the training data using Root Mean Squared Error (RMSE) and Structural Similarity Index Measure (SSIM) metrics.

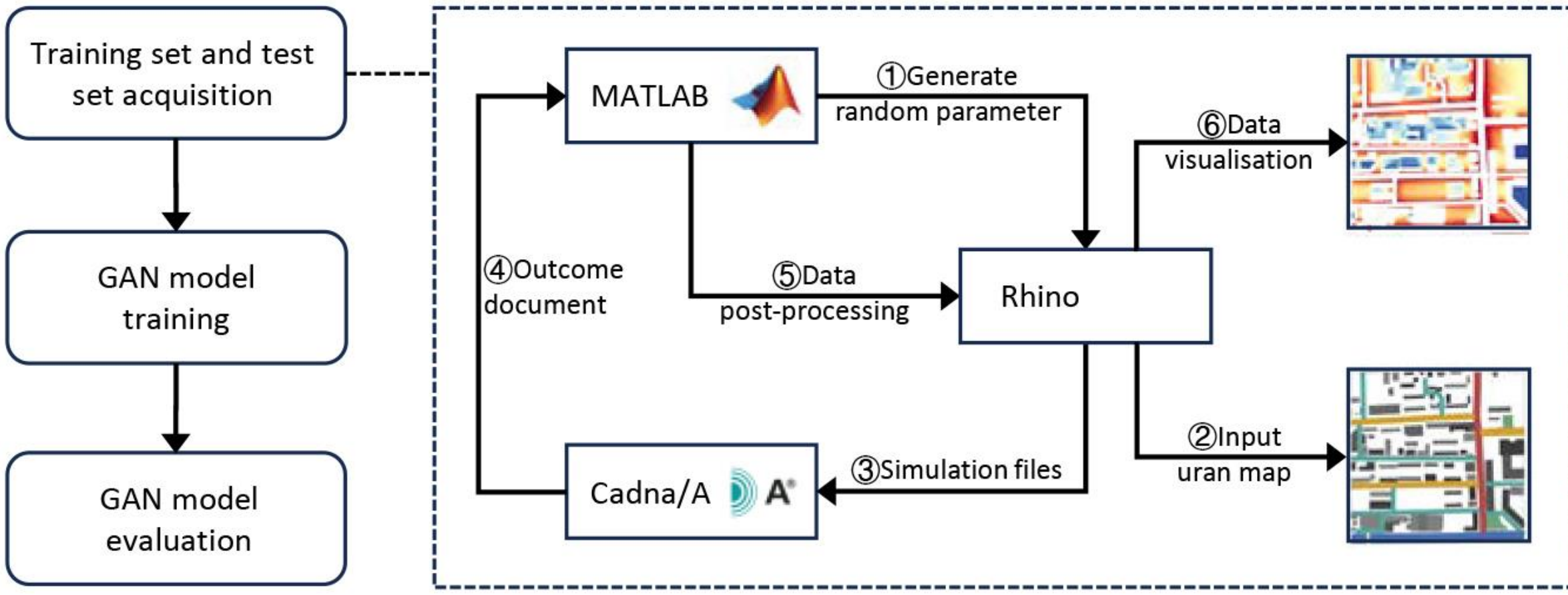

Figure 1 Framework of model establishment.

### 2.1 Dataset creation

To construct the dataset for this study, the data were generated through a series of steps designed to ensure that the model would be broadly generalisable and accurate, as illustrated in Figure 1.

**Step 1** Using the OpenStreetMap (OSM) open-source mapping platform, a representative urban area within a 5000 m by 2000 m range in Beijing, China, was selected. Key urban elements, including buildings, city roads, and green spaces, which substantially affect sound propagation and attenuation, were included.

**Step 2** MATLAB (version R2023a) was used to generate random parameters, following which the urban area was randomly cropped to 500 m by 500 m using Grasshopper. A total of 2,200 samples were thus produced, which served as inputs for the training set.

**Step 3** The simulation file, created by Grasshopper, was imported into Cadna/A (version 2018) environmental noise simulation software for acoustic modelling, simulating the spatial distribution of equivalent continuous A-weighted sound pressure levels ($L_{Aeq}$) generated by traffic conditions in each sample.

The calculation configuration is set in Cadna/A. The noise calculation uses the German RLS-90 (Richtlinien für den Lärmschutz an Strassen) noise prediction model, which follows the following principles (Quartieri et al., 2009): Ⅰ. only the first-order reflections; Ⅱ. no lateral diffraction at obstacles; Ⅲ. no meteorological correction; Ⅳ. the two outermost lanes of a road are calculated as separate line sources with a height of 0.5 m above the road.

Acoustic calculations are conducted on urban features such as buildings, green spaces, roads, and traffic flows. When simulating traffic noise, data concerning roadways and traffic volume typically reflect real-world conditions. Acquiring data for noise prediction at the urban level is challenging and involves intricate setup processes. The objective of this study is to offer a swift prediction of urban traffic noise during the initial design phases when design solutions are still in preliminary stages. Therefore, when configuring roads and traffic volume, we standardise this part of the content according to road classification. Design traffic information for roads of various classifications was established in accordance with pertinent Chinese road design specifications (Code for design of urban road engineering, CJJ37-2012; Technical Standard of Highway Engineering, JTC B01-2014) to simulate the traffic noise level under the most unfavourable conditions. The acoustic calculations for all the urban elements are specified in Table 1.

**Step 4** Upon completion of the noise simulation by Cadna/A software, the resulting simulation file with the .rst extension was imported back into MATLAB.

**Step 5** The ".rst" file contains positional information (x, y, z coordinates) and LAeq from simulation sampling points in Cadna/A. In MATLAB, the algorithm first discretizes the calculation area (500 m × 500 m) into a grid with each cell measuring 1 m × 1 m. The $L_{Aeq}$ from the simulation sampling points are subsequently mapped to the corresponding grid points on the basis of their positional information. If multiple simulation sampling points fall within a single grid cell, the $L_{Aeq}$ for that cell is computed as the average of these points. Finally, the algorithm employs cubic spline interpolation for further refinement. The processed results are then stored in a .csv file and transferred to Grasshopper for visualisation.

**Step 6** The visualisation of the data was executed via Grasshopper, with the resulting urban noise maps constituting the outputs for the training set.

Table 1 Acoustic calculation configuration

| | | | | | | | | | | |
|---|---|---|---|---|---|---|---|---|---|---|
| Building module | | Height (m) | 15-30 meters, randomly assigned | | | | | | | |
| | | Surface sound absorption coefficient | 0.2 (AG, 2006) | | | | | | | |
| Foliage module(China, 2021) | Propagation distance d (m) | Frequency (Hz) | 63 | 125 | 250 | 500 | 1000 | 2000 | 4000 | 8000 |
| | 10≤d＜20 | Attenuation /dB | 0 | 0 | 1 | 1 | 1 | 1 | 2 | 3 |
| | | Frequency (Hz) | 63 | 125 | 250 | 500 | 1000 | 2000 | 4000 | 8000 |
| | 20≤d＜200 | Attenuation （dB/m） | 0.02 | 0.03 | 0.04 | 0.05 | 0.06 | 0.08 | 0.09 | 0.12 |
| Roads and transportation module | | Road level | Urban freeway | | Main road | | Secondary road | | Branch road | |
| | | Number of lanes | Two-way six-lane | | Two-way eight-lane | | Two-way six-lane | | Two-way four-lane | |
| | | Road width (m) | 21 | | 28 | | 21 | | 14 | |
| | | Road surface type | Smooth mastix asphalt | | | | | | | |
| | | Road speed limit (small car, km/h) | 80 | | 60 | | 40 | | 30 | |
| | | Road speed limit (small car, km/h) | 60 | | 50 | | 30 | | 30 | |
| | | Traffic flow, vehicles/hour | 7400 | | 5700 | | 4350 | | 3200 | |
| | | Heavy truck ratio, % | 10 | | | | | | | |
| Ground absorption module | | Default value of 1.0 for ground absorption | | | | | | | | |
| Grid module | | Default grid spacing, 10 m; Grid height, 1.5 m | | | | | | | | |
| Meteorological module | | According to RLS-90, noise correction for meteorological conditions is ignored | | | | | | | | |

To prepare a database suitable for training the Pix2Pix model, all the results needed preprocessing. Each city plan and corresponding noise map were resised to 256×256 pixels. The database was then split, with 80% allocated for the training set and the remaining 20% allocated for the test set. Figure 2 illustrates a processed training case where Figure 2(a) displays the city plan, identifying buildings (with varying shades of grey indicating building heights from 15--30 m), roads (coloured blue for urban expressways, red for main roads, yellow for secondary roads, and lime green for branch roads), and urban green spaces (green). Figure 2(b) depicts the traffic noise map, with a colour gradient ranging from dark blue to dark red representing noise levels from 45 dB(A) to 80 dB(A)).

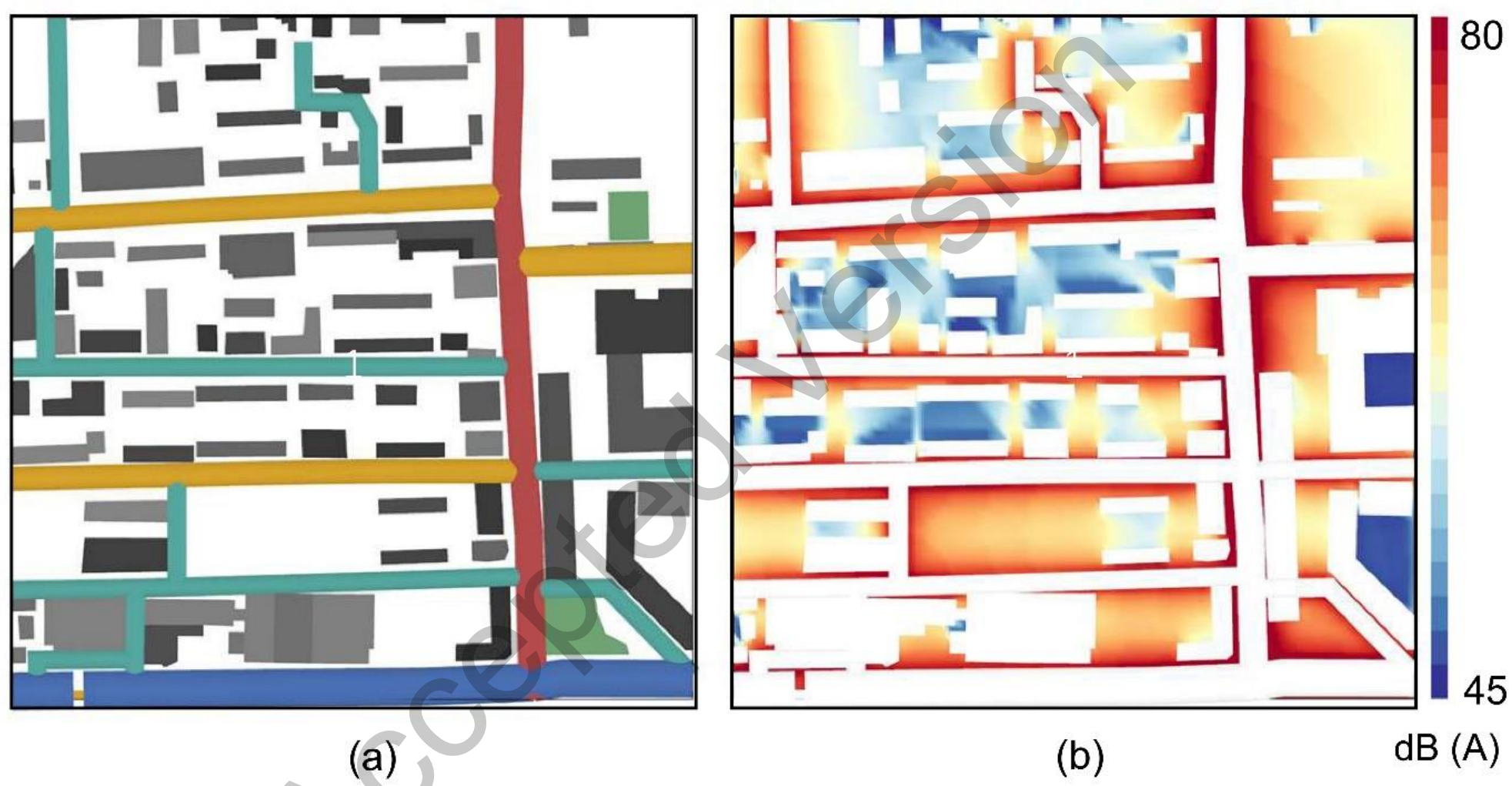

Figure 2 Processed training case: (a). the urban map; (b). the urban noise map.

**2.2 Pix2pix training**

The surrogate model was constructed using a GAN, specifically through the pix2pix algorithm. It comprises two components: a generator (G) and a discriminator (D). The generator G accepts a 256*256*3 urban map as its input and produces a 256*256*3 noise map as its output. The working principle of the algorithm is depicted in Figure 3.

The generator employs a U-net (Ronneberger et al., 2015) architecture that preserves low-level detail information at different resolutions through layer-hopping connections. To assess the authenticity of distinct image patches, Patchnet was used for the discriminator (Yang et al., 2019). The network structures of both the generator and the discriminator are shown in Figure 4.

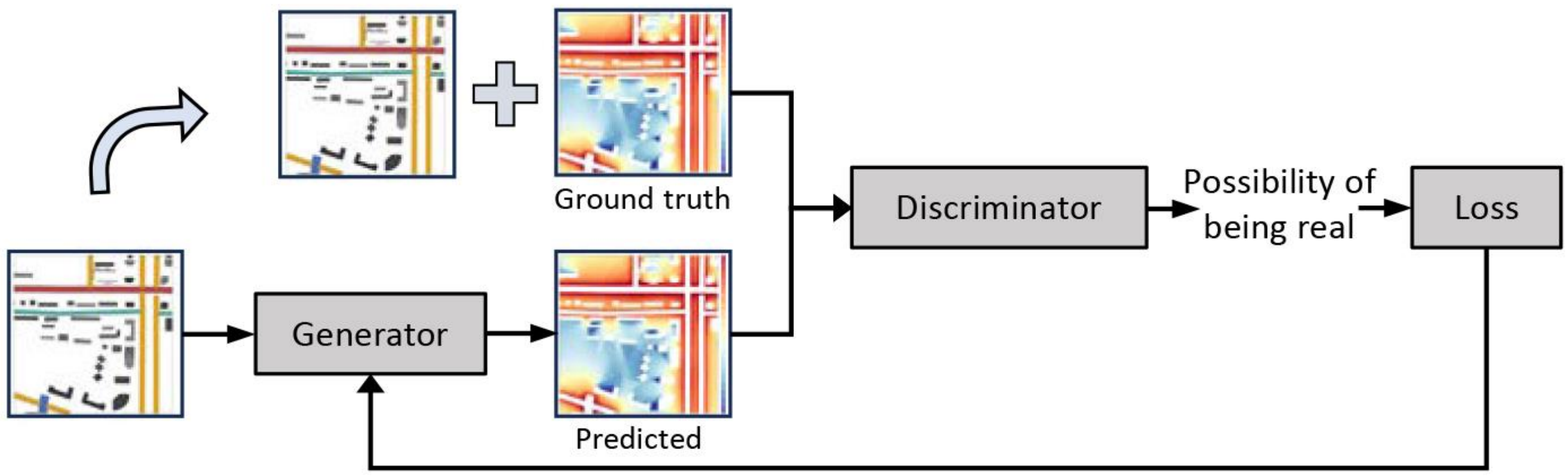

Figure 3 Principle of the pix2pix algorithm.

The optimisation objective (the loss function) in this model consists of two parts: a combination of adversarial loss and $L_1$ loss for $G$ and a binary cross-entropy loss function for $D$:

$$G^* = argmin(G)max(D)L_{cGAN}(G, D) + \lambda L_{L1}(G) \quad (1)$$

$$L_{cGAN}(G, D) = E_{x,y}[logD(x, y)] + E_{x,y}[\log(1 - D(x, G(x))] \quad (2)$$

$$L_{L1}(G) = E_{x,y}[||y - G(x)||_1] \quad (3)$$

where $G$ is the generator, $D$ is the discriminator, $x$ is the input urban map, $y$ is the ground truth urban noise map, $G(x)$ is the urban noise map generated by the generator, and $\lambda$ is the weight value.

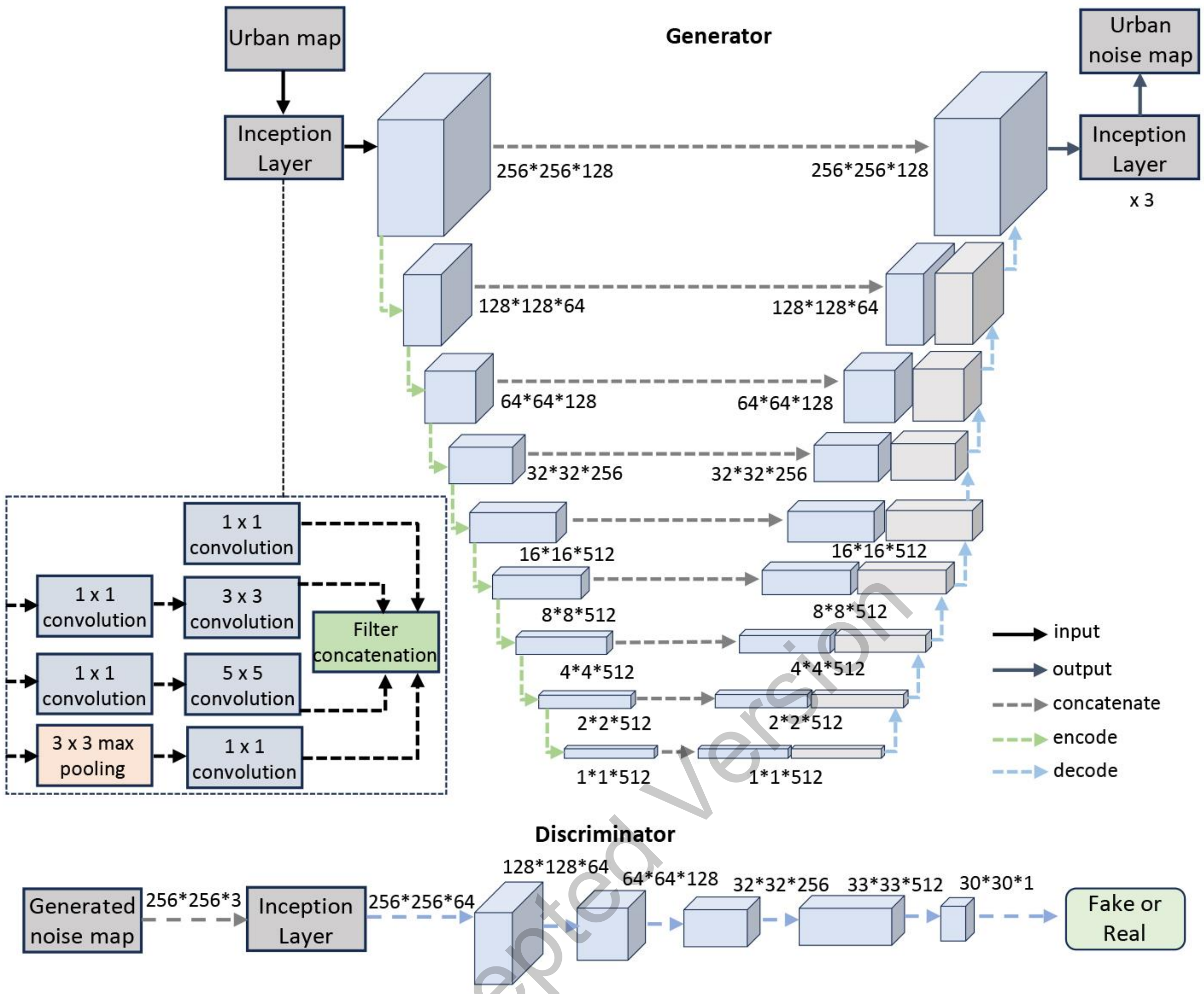

Figure 4 Network structure of the generator (G) and the discriminator (D).

## 2.3 Model accuracy

In this study, the model's performance is assessed using the accuracy metrics RMSE and SSIM. The RMSE gauges the average variance between the predicted values and actual values, as depicted in equation (4). A diminishing RMSE signifies minimal errors and heightened accuracy.

$$RMSE = \sqrt{\frac{1}{\mathrm{n}}\sum_{i=1}^{n}(y_i - \hat{y}_i)} \quad (4)$$

where $\mathrm{n}$ is the number of samples; $\hat{y}_i$ is the prediction value; and $y_i$ is the true value.

The SSIM serves to quantify and delineate the distinctions between the generated image and the source image (Wang et al., 2004). It amalgamates the assessment of image disparities grounded in the perception of image quality by the Human Visual System (HVS) and structural details. By assessing the luminance, contrast, and structure of two images, it mirrors the structural resemblance between them. The resultant SSIM index is a decimal value spanning from -1 to 1, where 1 signifies flawless similarity, 0 denotes no similarity, and -1 signifies complete anti-correlation. The computation of the SSIM is illustrated in formula (5).

$$SSIM(x,y) = \frac{(2\mu_x\mu_y+c_1)(\sigma_{xy}+c_2)}{(\mu_x^2+\mu_y^2+c_1)(\sigma_x^2+\sigma_y^2+c_2)} \tag{5}$$

where $\mu_x$ is the pixel sample mean of $x$; $\mu_y$ is the pixel sample mean of $y$; $\sigma_x^2$ is the variance of $x$; $\sigma_y^2$ is the variance of $y$; $\sigma_{xy}$ is the covariance of $x$ and $y$; and $c_1$ and $c_2$ are two constants used to stabilise the division with a weak denominator.

# 3 Results

As depicted in the learning loss curves presented in Figure 5, the discriminator converged after approximately 500 batches, whereas the generator reached convergence after approximately 1500 batches.

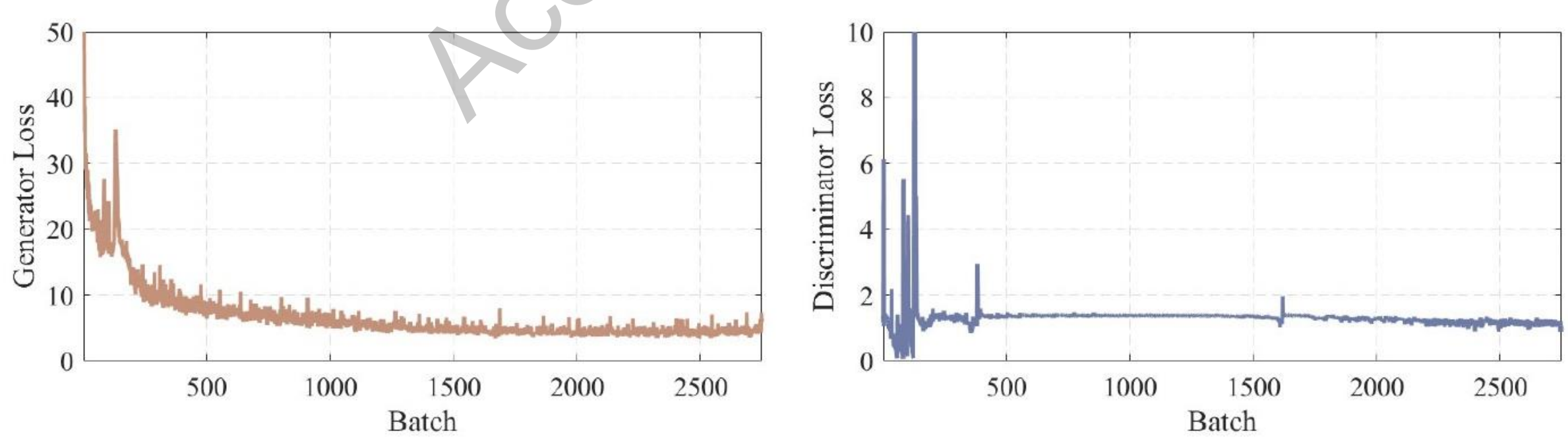

Figure 5 Losses of the generator and discriminator.

Table 2 shows the ground truth, predicted images, and grayscale difference images for selected cases within the validation dataset. Visually, the predicted images generated by the model and the ground truth convey nearly identical information regarding the distribution of traffic noise. The RMSE and the SSIM are also included in Table 2.

Table 2 Model performance evaluation (grayscale difference images, RMSE and SSIM)

| Input | Ground Truth | Predicted Result | Error Map | RMSE (dB(A)) | SSIM |
|---|---|---|---|---|---|
| 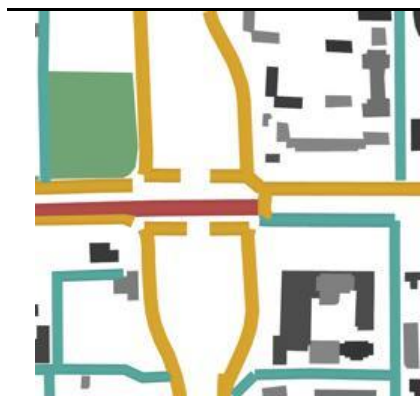 | 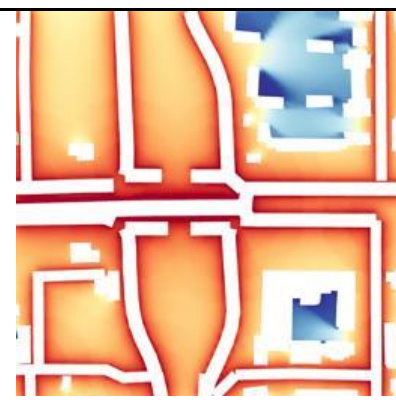 | 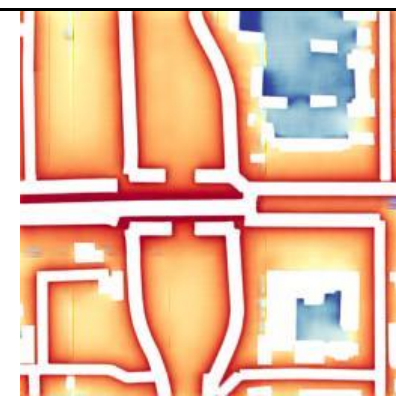 | 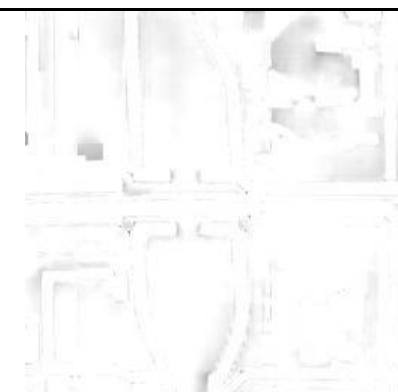 | 0.249 | 0.967 |
| 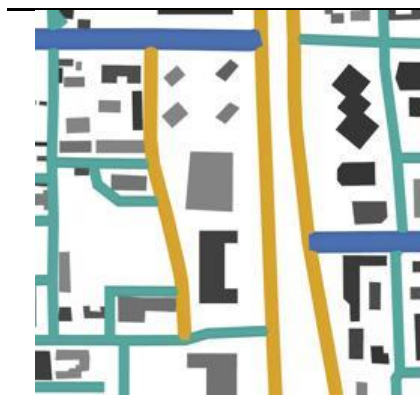 | 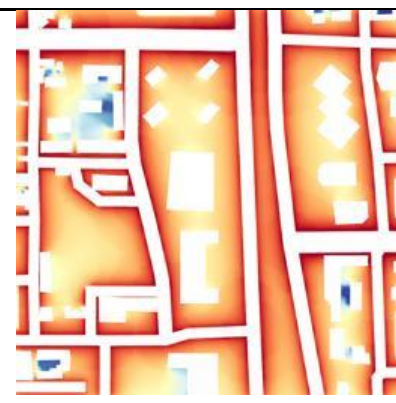 | 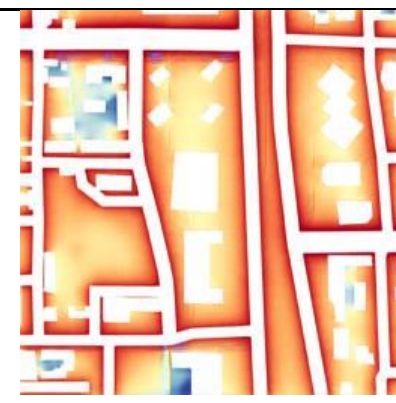 | 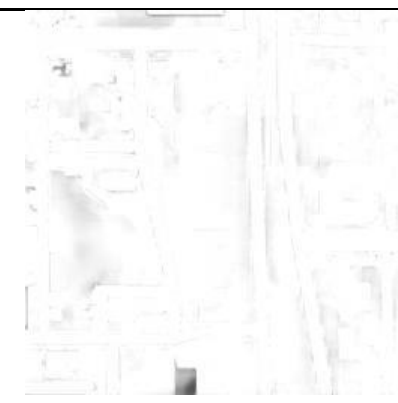 | 0.235 | 0.966 |
| 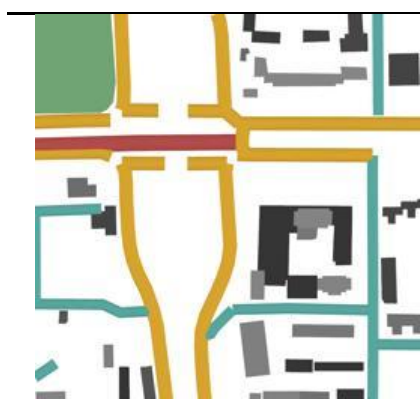 | 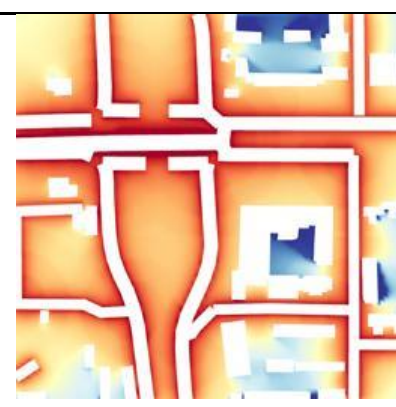 | 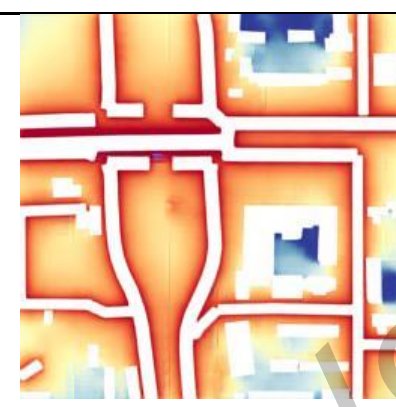 | 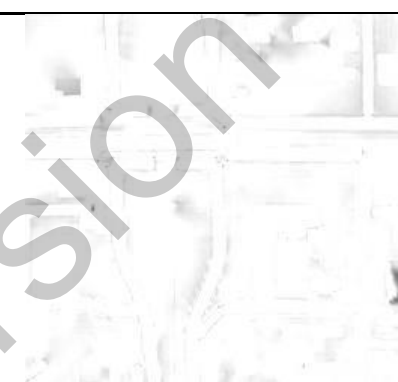 | 0.235 | 0.958 |
| 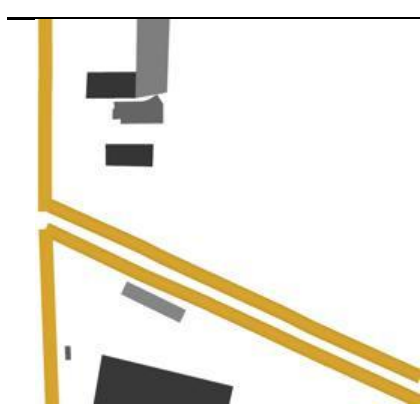 | 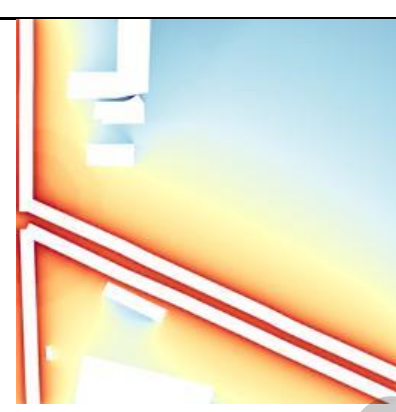 | 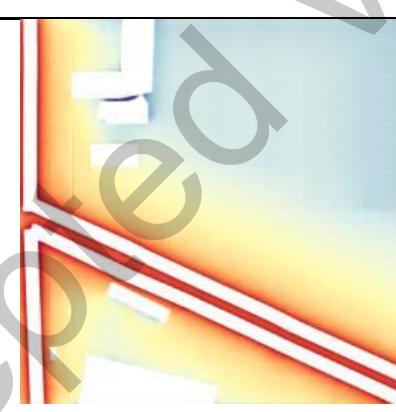 | 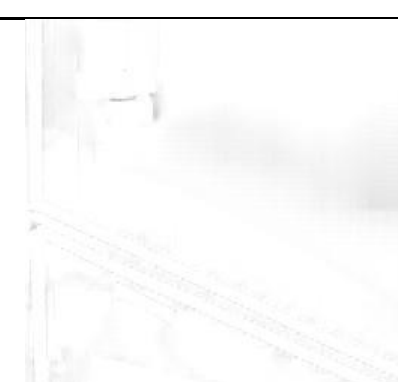 | 0.202 | 0.913 |
| 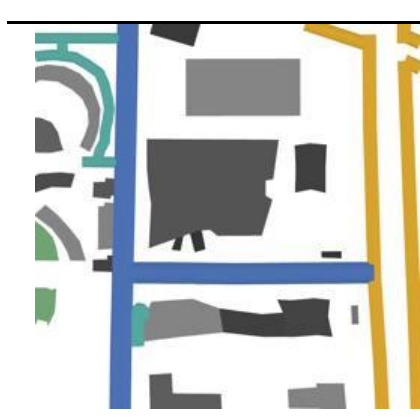 | 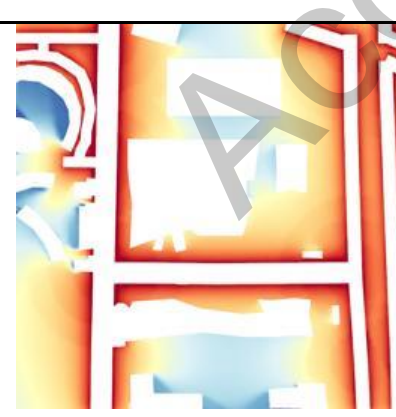 | 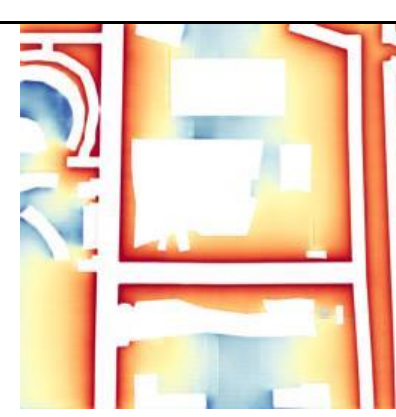 | 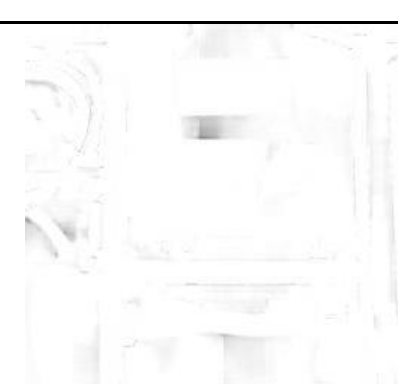 | 0.212 | 0.951 |
| 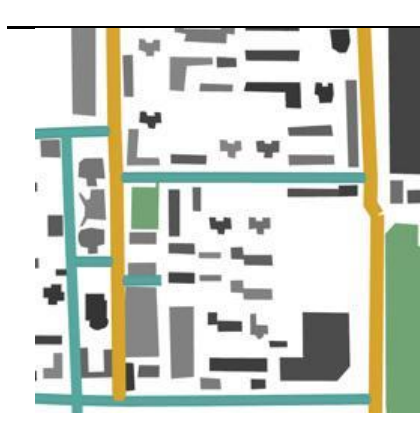 | 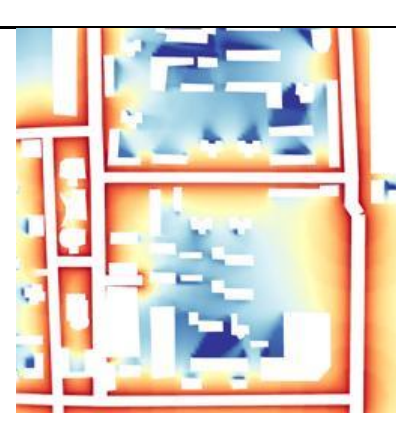 | 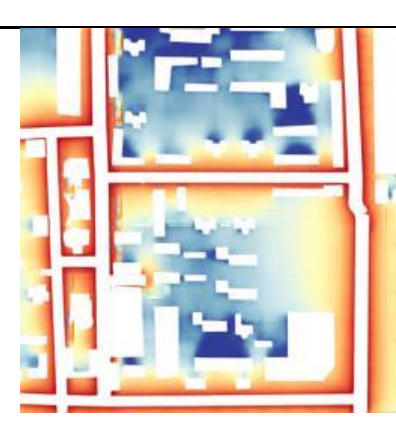 | 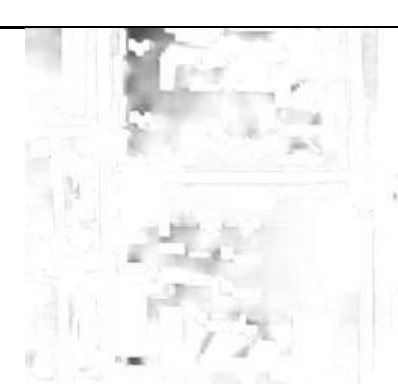 | 0.292 | 0.922 |

The trained surrogate model was applied to 440 validation datasets, and the RMSE and SSIM were computed. Figure 6 presents a histogram of the test set outcomes. The test set yields an average RMSE of 0.3024 dB(A) and an average SSIM of 0.8528. These results indicate that the surrogate model predictions closely align with the simulation software calculations. Consequently, the constructed model can offer decision-making support with similar accuracy to that of Cadna/A.

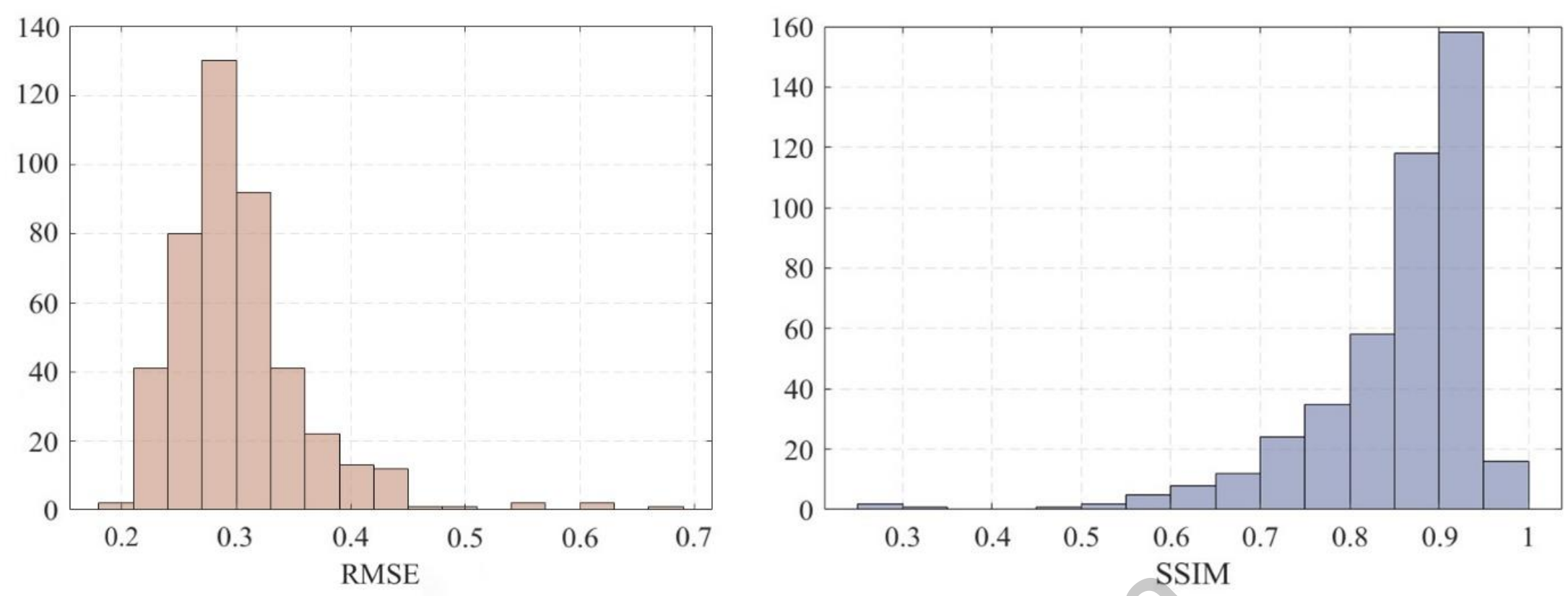

Figure 6 RMSEs and SSIMs of the validation dataset.

## 4 Integration

The trained surrogate models have been integrated into Rhino/Grasshopper, which serves as a readily usable tool to assist users in rapidly obtaining urban traffic noise maps during the initial phase of design. The tool comprises two principal modules: the model selection module and the fast prediction module.

The model selection module is configured in Grasshopper by connecting various component blocks. It captures the urban map sketched by the user in Rhino and transforms it into an image format compatible with the fast prediction module. This image includes elements such as buildings, road widths, and green spaces. Figure 7(a) shows the final city plan image utilised for predicting the urban traffic noise map.

Co-simulation with Grasshopper was performed by producing a program in MATLAB, which takes the build plan from the model selection module and calls the trained surrogate model. The resultant urban noise map prediction is displayed to the user within Grasshopper, as illustrated in Figure 7(b). Users wishing to refine the city layout may proceed to adjust the configurations of buildings, roads, and so forth in Rhino, enabling the generation of an optimised urban traffic noise map.

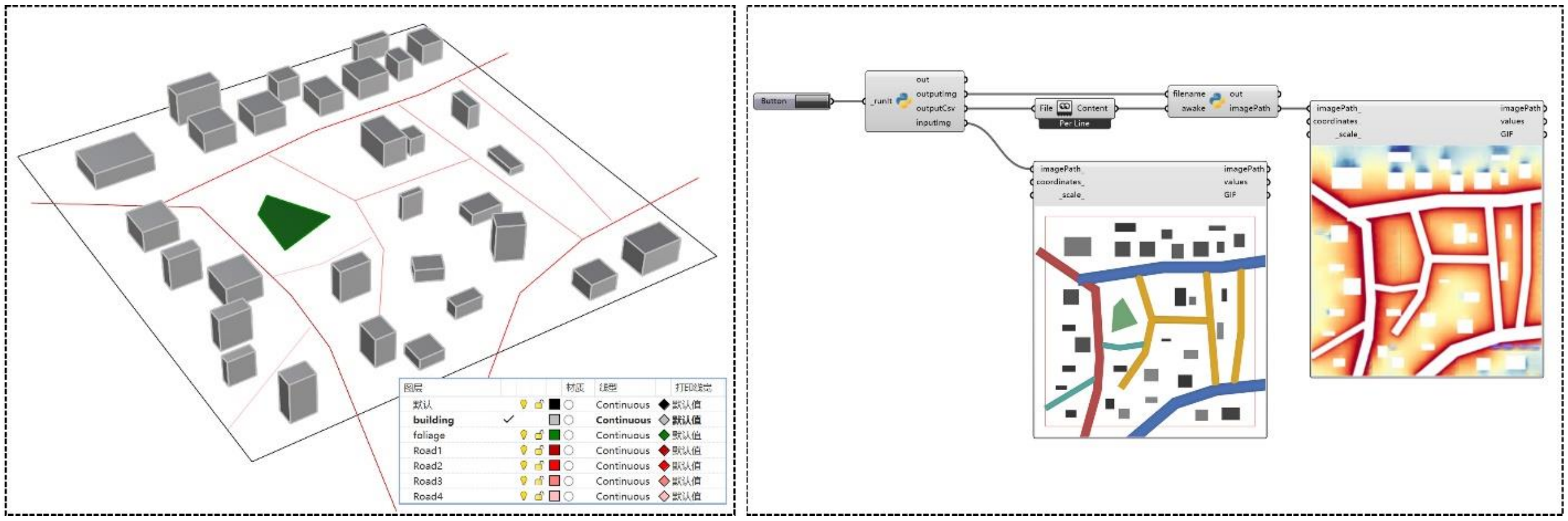

Figure 7 (a) Interfaces in Rhino; (b) Plugin interfaces in Grasshopper.

## 5 Conclusion and discussion

In this study, a GAN-based proxy model was constructed for the rapid prediction of urban noise maps and a design tool was developed based on this proxy model. The model's accuracy was evaluated by comparing the predicted results with those of the most widely used conventional noise mapping software, using the RMSE and SSIM as the metrics for assessment. The test set yielded mean values of 0.3024 dB(A) for the RMSE and 0.8528 for the SSIM, confirming the precision of the proxy model in predicting urban noise maps(Jiang et al., 2023; Kastner and Dogan, 2023; Li et al., 2024; Wan et al., 2023). These results demonstrate the viability of the GAN model for practical use.

The design tool, integrated within Rhino/Grasshopper, comprises two principal modules: the model selection module and the rapid prediction module. Designers can swiftly gauge the traffic noise conditions associated with preliminary urban planning scenarios based on the initial patterns of urban buildings and road networks. This enables them to implement optimised noise prevention and control designs during the early stages, thereby enhancing design efficiency.

Compared with traditional urban traffic noise prediction methods, the surrogate model proposed in this study provides operational convenience for incorporating noise considerations in the early design stages of urban planning. Traditional methods require the repeated construction of complex acoustic models when facing frequent changes in urban layouts and road design plans, which is time-consuming and computationally expensive. Thus, these methods fail to meet the rapid response demands of early design stages. In contrast, with the integration of design tools, the surrogate model developed in this study can instantly generate new noise prediction results following adjustments to the design schemes. This approach offers significant convenience for urban planners and road designers, particularly during the initial phases of a project, when continuous model optimisation and comparison are essential. In addition, this approach provides great convenience for non-acoustic experts, allowing them to skip complex acoustic modelling and acoustic procedures for noise prediction. To a significant extent, this can diminish the design input in the later stages of the project and avert financial expenditures for noise control post-completion.

The tool developed in this research aims to enable users to predict the distribution of traffic noise during the early design stages. It helps users analyse the traffic noise impact of the solution and assists in optimisation. Therefore, we focus solely on key design elements such as roads, buildings, and green spaces that affect noise generation and distribution (Guedes et al., 2011) while disregarding elements that cannot be determined in the early design stages and have minimal impact on acoustic calculations. In configuring key information for acoustic calculations, our focus was on standardising and unifying the processing of traffic information. Urban traffic conditions are intricate and encompass complex data configurations, including traffic volume, vehicle types and proportions, and temporal variations in traffic flow. During the initial design phases, solutions are typically preliminary and uncertain. Hence, to increase the usability of the tools developed in this study, we simplified and standardised the configuration settings associated with traffic. According to the Chinese Road Design Code, urban roads are classified into four types. For each road category, uniform settings for the road width, number of lanes, and traffic volume were established to simulate the least favourable scenario. Despite this, this study emphasises the feasibility and practicality of using a GAN to develop surrogate models for the rapid prediction of urban traffic noise. In future research, we will further expand the functionality of this tool on the premise of achieving rapid prediction capabilities. The enhancements include refining the configuration of road traffic-related information, adding ports for user input of real-time traffic data and other computational configurations, and integrating predictive modules with active noise reduction design (such as sound barriers) to enhance the tool's applicability in traffic noise assessment throughout the entire design phase. In addition, in terms of noise impact, different indicators and time spans of noise effects should not be overlooked (Botteldooren, 2023). Therefore, in addition to considering the inclusion of multimodal variable types to increase the predictive accuracy of models, future research should incorporate more noise indicators and considerations regarding time dimensions. This will further address various related needs, including the evaluation of noise reduction measures, the analysis of noise risks to sensitive buildings and exposed populations, urban noise management, and the management of sound environment functional areas.

Although the model demonstrates superior predictive ability in terms of accuracy and image similarity, it fails to accurately predict all samples (Table 3). Table 3 lists the prediction samples with significant errors, where the RMSE values range from 0.454 dB(A) to 0.675 dB(A) and the SSIM values range from 0.252 to 0.696. Compared with Table 2, their prediction accuracy has markedly decreased. Upon observation, most of these samples are large urban open spaces or green spaces, and the limited number of such training samples may account for the poor results. The study selected a high-density city, Beijing, for urban map collection, characterised by a dense distribution of buildings and urban roads. In Table III, the slightly denser samples of buildings and roads (corresponding to SSIM values of 0.696 and 0.690) still exhibit better predictive performance than the large open space or green space samples do. This further indicates that the model more accurately learns the noise distribution patterns of buildings and roads. In the future, the sample sizes of more types of typical urban spaces should be increased to fully understand the spatial patterns of urban open spaces and large green areas, thereby further improving the prediction accuracy.

Table 3 Low-quality model performance evaluation

| Input | Ground Truth | Predicted Result | Error Map | RMSE (dB(A)) | SSIM |
|---|---|---|---|---|---|
| 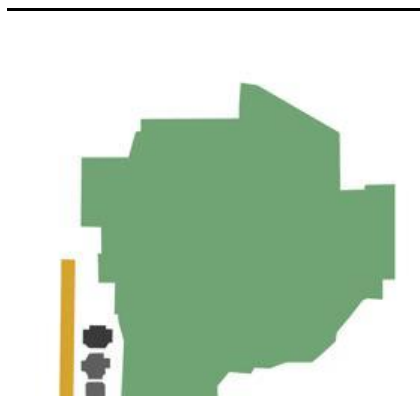 | 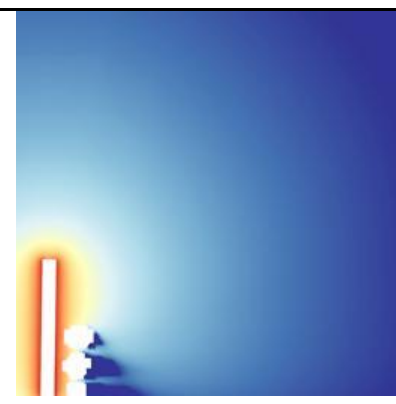 | 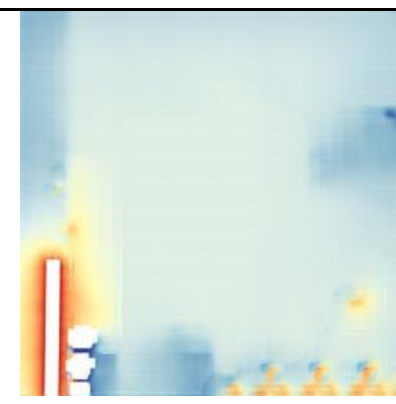 | 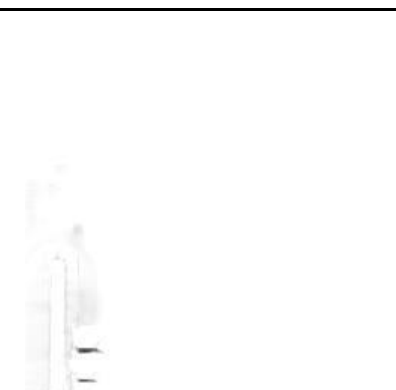 | 0.609 | 0.314 |
| 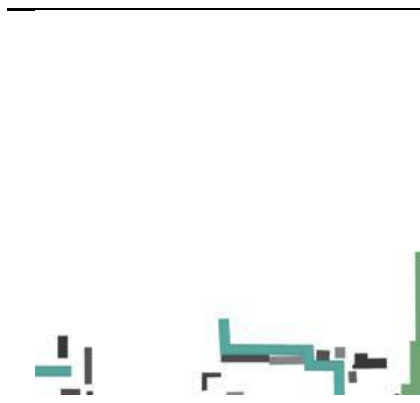 | 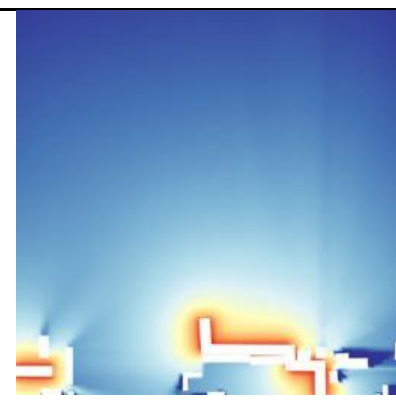 | 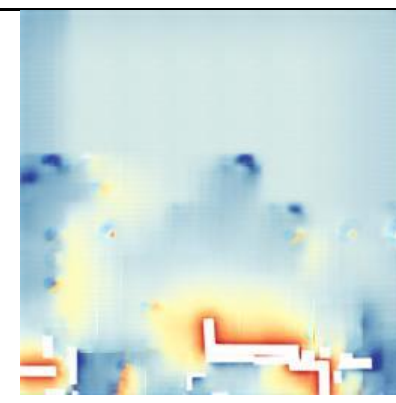 | 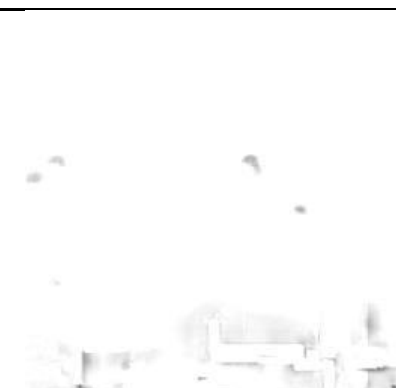 | 0.552 | 0.468 |
| 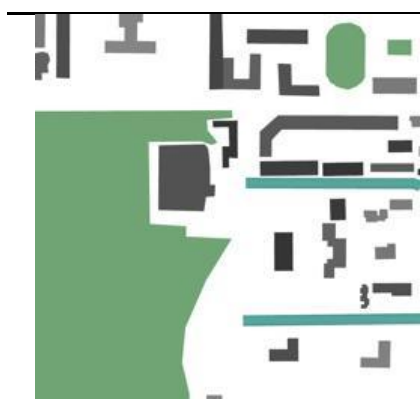 | 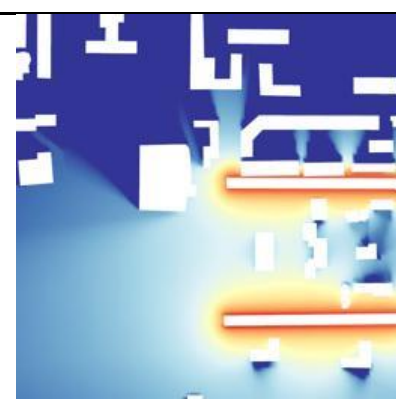 | 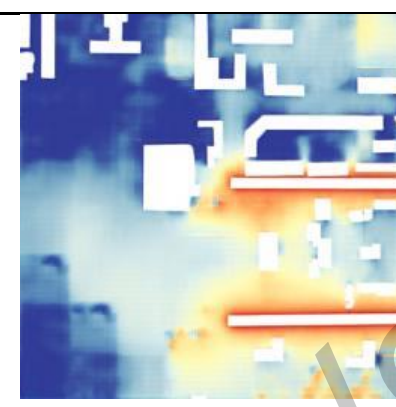 | 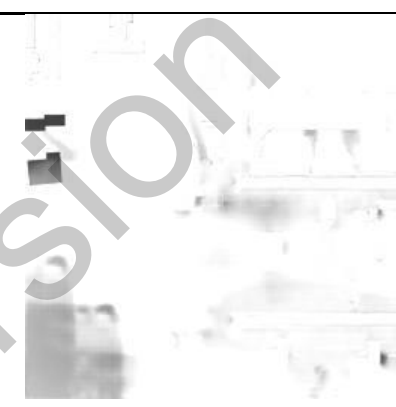 | 0.454 | 0.696 |
| 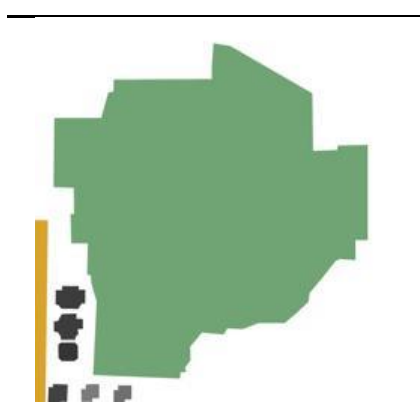 | 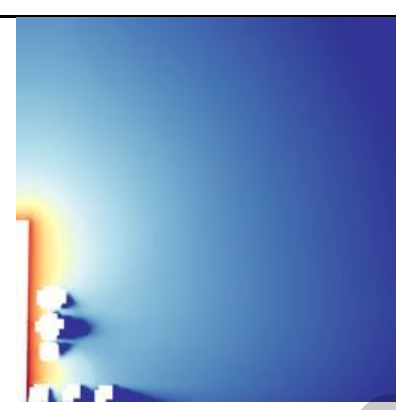 | 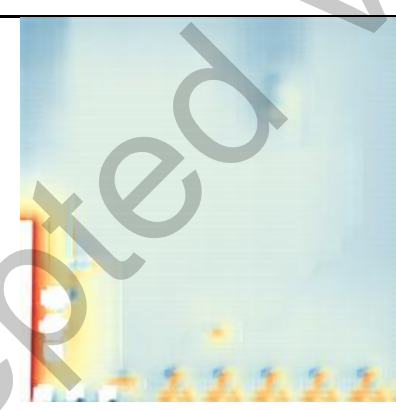 |  | 0.618 | 0.250 |
| 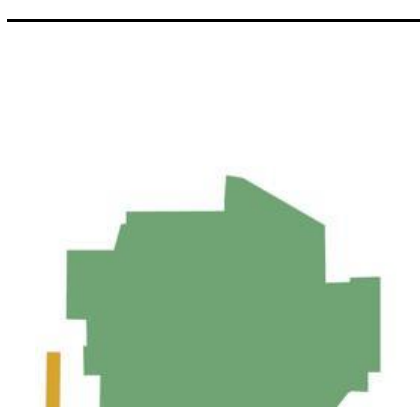 | 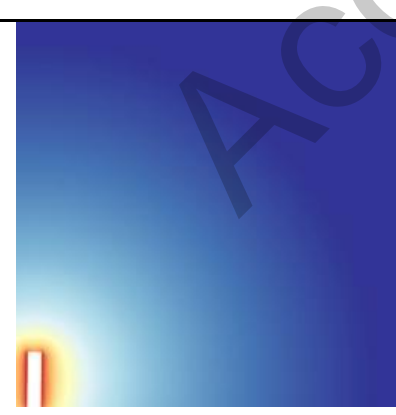 | 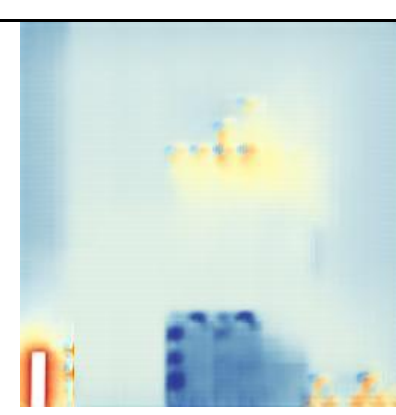 | 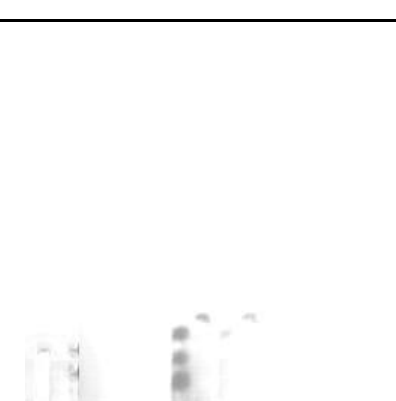 | 0.675 | 0.252 |
| 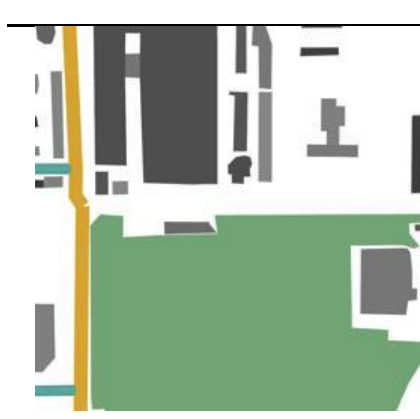 | 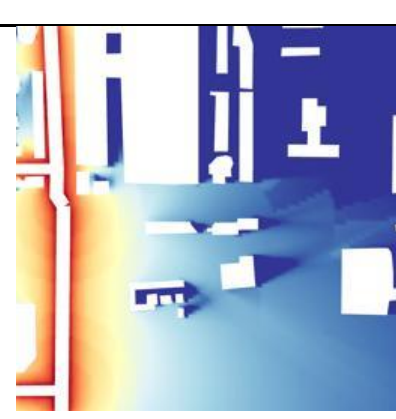 | 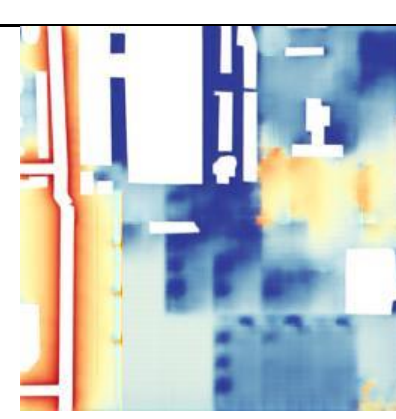 | 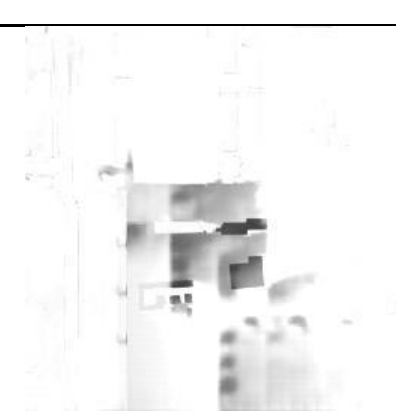 | 0.497 | 0.690 |

In summary, the method developed in this research not only offers advantages in terms of efficiency and flexibility, enabling rapid iterations of design schemes but also provides a highly efficient and convenient tool while maintaining prediction accuracy. As a design tool, this approach also reduces the barrier to entry for non-acoustic experts while enabling quick predictions to facilitate subsequent parametric optimisation design linkage. This significantly supports practical applications in urban planning, road design, and environmental impact assessments. It is important to note that as it currently stands, the surrogate model developed in this study is not intended to replace traditional noise mapping techniques, but rather to supplement them, particularly during the early design stages and for use by non-experts. While this model offers significant advantages in terms of efficiency and flexibility, traditional methods remain more universally applicable, and should still be employed for detailed noise assessments and final design evaluations.

## Author contribution

Xinhao Yang: Methodology, Investigation, Writing – original draft; Zhen Han: Conceptualization, Methodology, Investigation; Xiaodong Lu: Writing - review & editing; Yuan Zhang: Conceptualization, Writing - review & editing, Supervision, Funding acquisition. Xinhao Yang and Zhen Han contributed equally to this work.

## Declaration of competing interest

Yuan Zhang, Xinhao Yang, and Zhen Han have a patent #202410572844.7 under review related to the subject matter of this article. This patent could benefit financially from the publication of this research. The authors declare that this interest did not influence the study's design, data collection, analysis, or interpretation. Except for this, the authors have no other known competing financial interests or personal relationships that could have appeared to influence the work reported in this paper.

## Data availability

The dataset used/generated in this study is available from the corresponding author upon reasonable request for academic use.

## Code availability

The code used in this study to implement the general pix2pix model (https://www.tensorflow.org/tutorials/generative/pix2pix?hl=zh-cn) is written in MATLAB.

The code is available upon reasonable request from the corresponding author for academic use.

## Acknowledgement

Y.Z. acknowledges the National Key Research and Development Program of China (No. 2023YFC3804102-04) and the Educational Department of Liaoning Province Fundamental Research Project (No. JYTZD2023172).